\documentclass[twocolumn,times,final,authoryear]{elsarticle}
\pdfoutput=1

\usepackage{prletters}
\usepackage{framed,multirow}

\usepackage{amssymb}
\usepackage{latexsym}
\usepackage{graphicx}
\usepackage{amsmath}
\usepackage{amssymb}
\usepackage{booktabs}

\usepackage{url}
\usepackage{xcolor}
\definecolor{newcolor}{rgb}{.8,.349,.1}

\journal{Pattern Recognition Letters}

\newcommand\blfootnote[1]{%
  \begingroup
  \renewcommand\thefootnote{}\footnote{#1}%
  \addtocounter{footnote}{-1}%
  \endgroup
}

\begin{document}

\setcounter{page}{1}

\begin{frontmatter}

\title{Few Shot Class Incremental Learning using Vision-Language models}

\author[1,4,*]{Anurag  \surname{Kumar}${}^\dagger$}

\author[1,5,*]{Chinmay \surname{Bharti}${}^\dagger$}

\author[2,*]{Saikat \surname{Dutta}\corref{cor1}}
\cortext[cor1]{Corresponding author}
\ead{23d2031@iitb.ac.in}

\author[3]{Srikrishna  \surname{Karanam}}

\author[1]{Biplab  \surname{Banerjee}}


\affiliation[1]{organization={IIT Bombay},
                city={Mumbai}, 
                country={India}}

\affiliation[2]{organization={IITB-Monash Research Academy},
                city={Mumbai}, 
                country={India}}

\affiliation[3]{organization={Adobe Research},
                city={Bengaluru}, 
                country={India}}

\affiliation[4]{organization={
Intercontinental Exchange Inc},
                city={Hyderabad}, 
                country={India}}

\affiliation[5]{organization={CodeAnt AI},
                city={Bengaluru}, 
                country={India}}

\received{1 May 2013}
\finalform{10 May 2013}
\accepted{13 May 2013}
\availableonline{15 May 2013}
\communicated{S. Sarkar}

\begin{abstract}
Recent advancements in deep learning have demonstrated remarkable performance comparable to human capabilities across various supervised computer vision tasks. However, the prevalent assumption of having an extensive pool of training data encompassing all classes prior to model training often diverges from real-world scenarios, where limited data availability for novel classes is the norm. The challenge emerges in seamlessly integrating new classes with few samples into the training data, demanding the model to adeptly accommodate these additions without compromising its performance on base classes. To address this exigency, the research community has introduced several solutions under the realm of few-shot class incremental learning (FSCIL). 

In this study, we introduce an innovative FSCIL framework that utilizes language regularizer and subspace regularizer. During base training, the language regularizer helps incorporate semantic information extracted from a Vision-Language model. The subspace regularizer helps in facilitating the model's acquisition of nuanced connections between image and text semantics inherent to base classes during incremental training. Our proposed framework not only empowers the model to embrace novel classes with limited data, but also ensures the preservation of performance on base classes. To substantiate the efficacy of our approach, we conduct comprehensive experiments on three distinct FSCIL benchmarks, where our framework attains state-of-the-art performance.
\end{abstract}


\end{frontmatter}

\blfootnote{$^1$ Equal Contribution}
\blfootnote{$^\dagger$ Work done at IIT Bombay}


\section{Introduction}


Given the non-stationary nature of data generation in many application areas, it becomes crucial for learning to occur continuously over time. Incremental learning addresses this challenge by enabling models to adapt continually to new and non-overlapping tasks, while ensuring the maximum retention of knowledge from previous tasks to facilitate real-time inference.
Significant advancements have recently been achieved, particularly in the field of class incremental learning (CIL). Typically, a CIL model \citep{belouadah2019il2m, zhu2021class, van2021class} is trained on a fresh set of categories using a substantial volume of training samples during each session. However, this practice inevitably leads to parameter drift, causing the model to forget previously learned concepts from the previous set of classes. This phenomenon is known as catastrophic forgetting. Unfortunately, obtaining a sufficient number of training samples is not always feasible, especially for rarer and less frequent categories. Moreover, accessing data from previous classes may be restricted due to privacy concerns or limited memory availability for storing previous-class data.

These challenges give rise to the problem of few-shot class incremental learning (FSCIL). In FSCIL, abundant supervision is available for the initial base classes, while only a few training samples are accessible for the new classes introduced in subsequent learning sessions. FSCIL is notably more demanding than CIL because the limited amount of available data makes the model susceptible to overfitting. Additionally, the unavailability of data from previous sessions during a given training stage impacts the model's stability. 


Prior works on FSCIL have used nonstandard prediction architectures \citep{tao2020few_topic} or optimizing non-likelihood objectives \citep{ren2019incremental, yoon2020xtarnet}. Thus, state-of-the-art FSCIL methods often require nested optimizers, complicated data structures, and increased number of hyperparameters. How to apply representation learning and optimization techniques developed for standard classification problems to the incremental setting still remains an underexplored direction. Specifically, how one can connect class representations with semantic information in FSCIL needs further research. On the other hand, despite the recent success of Vision-Language models (VLMs) on various tasks \citep{clip,openseed}, state-of-the-art FSCIL methods are yet to utilize their usefulness. 

\textbf{Contributions}
Motivated by the aforesaid discussions, we propose a novel scheme where we make use of language regularizer and  subspace regularizer with additional image-to-text based relational model. In this approach, first we improve base model training using language regualizer which incorporates semantic information obtained from a Vision-Language model. We leverage an existing architecture based on the concept of subspace regularizers with the intention to solve the problems of overfitting and catastrophic forgetting. We modify the subspace regularizers used to span the base classes for developing the ability to interlink information between image and text semantics.

Our notable contributions are as follows:
\begin{itemize}
\item Our approach bridges the domain gap of text and image semantics in the base model in FSCIL.
\item We proposed a methodology for prompt engineering of the CLIP language models which can be further used for zero-shot and few-shot learning related works.
\item Our proposed approach beat the state-of-the-art methods by around $7\%$ for the CIFAR-FS dataset and by around $1\%$ for Mini-Imagenet and Tiered-Imagenet datasets.
\end{itemize}

\vspace{-15px}
\section{Background and Related Works}




Incremental learning focuses on building artificially intelligent
systems that can continuously learn to adapt to new tasks
from new data while preserving knowledge learned from previously learned tasks. 
One specific type of incremental learning is Class incremental learning (CIL), where unseen class examples arrive sequentially and the model is required to learn to classify all the classes incrementally. In each training session of CIL, the model has only access to the data for the current session. Sometimes, a small memory is considered that can be used to store some exemplars from previous training sessions \citep{masana2022class, zhou2023deep}. 

\textbf{Few-shot class incremental learning } Few-shot class incremental learning (FSCIL) is a learning approach that aims to progressively acquire knowledge about new classes with minimal supervision, while ensuring that previously learned classes are not forgotten. The first work on FSCIL is TOPIC \citep{tao2020few_topic}, which constructs and preserves the feature manifold topology formed by different classes. However, TOPIC faces a significant network capacity problem. Another method called LEC-Net \citep{lecnet} utilizes a self-activation module for dynamic expansion and compression of network nodes. However, the initial expansion phase in LEC-Net leads to overfitting, resulting in a sharp decline in performance.

On the other hand, FSLL \citep{fsll} is a regularization-based approach that handles overfitting and catastrophic forgetting by training only a subset of all the model parameters in each session rather than training the entire model. This helps mitigate the negative impact on performance. More recently, F2M \citep{f2m} proposed a solution to FSCIL by searching for flat local minima of the base training objective function and subsequently fine-tuning the model parameters within that flat region. \cite{fsil_gan} use a memory replay-based approach for FSCIL, but in place of traditional image replay, feature replay is utilized. To deal with the privacy concerns of vanilla data
replay, \cite{data_free} propose the data-free replay scheme for synthesizing old samples in FSCIL setting.
\cite{metafscil} have introduced meta-learning approach
for FSCIL. 


\textbf{Learning class representations}
In recent times, several methods have focused on the realm of class representations within few-shot and incremental learning scenarios \citep{tao2020few_topic}. \cite{imprinted} initialize novel class representations by utilizing the average features extracted from few-shot samples. In another class of approaches \citep{lwof,yoon2020xtarnet,zhang2021few}, a class representation predictor is trained using meta-learning. \cite{tao2020few_topic} enforce topological constraints on the manifold of class representations as new representations are incorporated. On the other hand, \cite{chen2020incremental} model the visual feature space as a Gaussian mixture and the cluster centers are utilized in a similarity-based classification scheme. Some other approaches introduce auxiliary schemes to condition both old and new class representations in each session e.g. \cite{zhang2021few} employ a graph attention network, while \cite{zhu2021self} utilize relation projection.

In relation to our approach, we draw connections to the works of  \cite{ren2019incremental} and \cite{akyurek2021subspace}. \cite{ren2019incremental}  proposed a nested optimization framework to learn auxiliary parameters for each base and novel class, aiming to influence the novel weights through regularization. \cite{akyurek2021subspace} demonstrate that these regularization targets can be geometrically derived without the necessity of an inner optimization step. Their regularization approach enables the learning of class representations that extend beyond strict linear combinations of base classes \citep{barzilai2015convex}. 

\textbf{Making use of language models}
By leveraging the inherent semantic representations learned from vast datasets, language models can grasp the relationships between words, enabling them to infer connections between unseen concepts.
This is especially relevant in zero-shot learning and generalized zero-shot learning, where supplementary information serves as the sole resource for understanding unfamiliar classes \citep{chang2008importance, larochelle2008zero, akata2013label, pourpanah2022review}.
In the realm of few-shot learning, specialized approaches have been developed to effectively incorporate auxiliary information. \cite{schwartz2022baby} use a combination of multiple semantic information such as category labels, attributes and natural language descriptions for few-shot image classification. 
\cite{cheraghian2021semantic} proposed a knowledge distillation approach for FSCIL which utilizes word embeddings for class labels. Traditional techniques employed in zero-shot and few-shot learning offer numerous solutions to mitigate overfitting and enable adaptability when encountering limited data availability.

\begin{figure*}[!htb]
    \centering
    \includegraphics[width=0.7\textwidth]{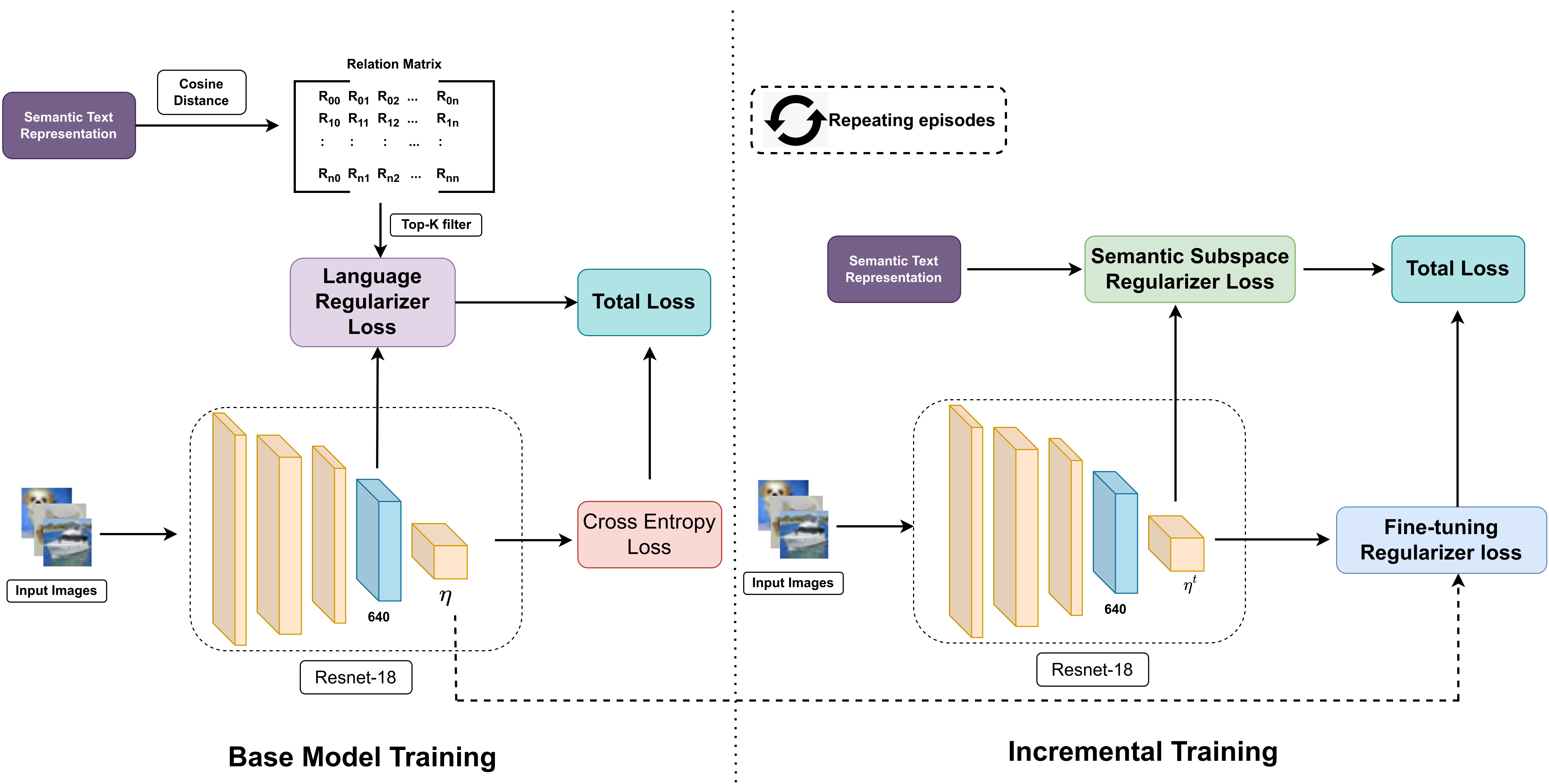}
    \vspace{-5px}
    \caption{Overview of our few-shot incremental learning framework. (a) For base model training, we use cross-entropy loss and language regularizer loss. (b) For incremental training, Semantic Subspace Regularizer loss and fine-tuning regularizer loss are used. Weight regularization for backbone and classifier is ignored in the diagram for simplicity.}
    \vspace{-10px}
\end{figure*}

\vspace{-10px}
\section{Problem Definition}
Let us begin with the formulation of the FSCIL Setup. We deal with a stream of T learning sessions each associated by the subset of the dataset denoted by $D^{(0)}$, $D^{(1)}$,..., $D^{(T)}$. Every subset $D^{(t)}$ is further divided into two sets called \textbf{support set} $S^{(t)}$ and \textbf{query set} $Q^{(t)}$. $S^{(t)}$ is used for training and $Q^{(t)}$ is used for evaluation during every learning session. A major portion of the dataset is allocated to $D^{(0)}$ which forms the base class set and the remaining portion is allocated uniformly to form the novel class sets.

We denote the classes for a set of examples S as $C(S) = \{y : (x, y) \in S\}$. The problem setup we have developed is incremental i.e, each support set comprises of just novel classes $( C^{(t)} \cap C^{(<t)}= \phi )$ whereas each query set evaluates models on a hybrid test set of novel and previously exposed classes $( C(Q^{(t)})=C^{(\leq t)} )$.
Further, we define our setting as few-shot in the set that every incremental set $|S^{(t)}|$ is considerably small than the set of base classes $|S^{(0)}|$.

At every incremental stage $t>0$, the aim is to fine-tune the existing classifier to accommodate information from the novel classes without compromising with ability for classification of previously seen classes.

\vspace{-10px}
\section{Methodology}
\vspace{-5px}
\subsection{Approach Overview}
Our approach to the problem statement can be broadly classified into two parts: base session and the incremental session. In the base session, we do a joint training of feature extractor and classification layer with the language model layer interacting between the two. The language model layer exposes the text semantics to the visual training process making the base model robust in both domains.

We use alternate training to introduce text via the language layer model. For the first $100$ epochs of base training, we use a loss function comprising of feature extractor and the classification layer. In the next $100$ epochs, we invoke the language model layer to introduce text semantics in the visual model.

In the incremental session, we freeze the base backbone and modify only the classification layer as per the information received from the stream of novel classes.

\subsection{Base Model Training}
We begin by implementing the training schema for base classes. The alternate training process can be mathematically represented as follows:
\begin{equation}
 \textit{L}_{base}= e_{1}.\textit{L}_{m}+ e_{2}.\textit{L}_{lreg}
\end{equation}

where,
\[
e_{1} = I (1\leq epoch \leq 100)
\]
\[
e_{2} = I (101\leq epoch \leq 200)
\]
Here, $I(.)$ is the indicator function. $L_m$ is used to train the backbone and classifier and $L_{lreg}$ is a regularization term that acts as a language regularize. $L_m$ is given by,

\begin{multline}
L_{m}(\eta,\theta)=\frac{1}{|S^{(0)}|} \sum_{(x,y)\in S^{(0)}} log  \frac{ exp(\eta^{T}_{y} f_{\theta}(x))       }{ \sum_{c\in C^{(0)}} exp(\eta^{T}_{c} f_{\theta}(x))} \\
 - \alpha(||\eta||^{2} + ||\theta||^{2} ) 
\end{multline}

Here, $f_{\theta} $ is the non-linear backbone feature extractor and weights of the last linear layer are denoted by $\eta$.  $f_{\theta} $ is implemented as the 
convolutional neural network based on ResNet-18.

The language model layer is designed to bridge the knowledge gap between image and text semantics. We initiate our approach by employing a technique called the cross-domain graph Laplacian regularization \citep{gune2018zero}. Its purpose is to preserve the underlying class structure of both semantic text representations and latent visual representations. To achieve this, we establish a neighborhood relationship using a graph representation based on K-nearest neighbor (K-NN) classes. This K-NN graph is constructed within the semantic space, with the class prototypes serving as nodes. An edge is present between two nodes if they are among the top K-nearest neighbors of each other, determined by the Euclidean distance in the semantic space. We have followed a simple binary 0-1 scheme for assigning edge weights. Mathematically, the K-NN graph can be represented by a similarity or adjacency matrix denoted as $S$.

\begin{equation}
  R_{ij} =
    \begin{cases}
      1 & \text{if $i$ and $j$ are top-$K$ neighbors}\\
      0 & \text{otherwise}
    \end{cases}       
\end{equation}


Here $R \in \{0,1\}^{s\times s}$ is used to define the neighborhood class structure in the semantic space, which is formed using the aforementioned approach. The process for selecting only the top K nearest neighbors comes from the idea of regularization of the effect of language in the training process. 
In the $R$ matrix, $R_{i}$ row contains the value 1 at positions that are in the top K of the Euclidean distance to them in the textual domain.
For one row $i$:
\begin{equation}
\sum_{j=1}^{N} R_{ij} = K
\end{equation}

Next, we compute cosine similarity between text embeddings of each pair of classes.
The cosine similarity between class $i$ and classes $j$ is given by,
\begin{equation}
\lambda _{ij} = \frac{ l_{i} .  l_{j} }{ \| l_i\| \| l_j\|}  
\end{equation}
where $l_i$ and $l_j$ are the text embeddings for the $i^{th}$ and $j^{th}$ class respectively. Similarly, cosine similarity between visual embeddings between class $i$ and class $j$ is given by, 
\begin{equation}
    \mu_{ij} = \frac{ v_{i} .  v_{j} }{ \| v_i\| \| v_j\|}
\end{equation}
where $v_i$ and $v_j$ are average visual representation for the $i^{th}$ and $j^{th}$ class respectively.

The top-K neighbors for every $i$-th class are grouped together
and then to align the neighborhood class structures of both the semantic and latent visual representations, we employ the graph Laplacian loss, defined as follows: 
\begin{equation}
L_{GRL}=\sum_{ (x,y) \in S^{(0)}} \sum_{k : R_{ky} = 1} |\lambda_{ky} - \mu_{ky}|
\end{equation}


Please note, we can use any other similarity  or distance measure apart from cosine similarity for computing $\lambda$ and $\mu$.
This loss function, $L_{GRL}$ forms the $L_{lreg}$ component of $L_{base}$.
During training, we typically take the value of $K$ as $5\%$ of the total class volume of the dataset.

\vspace{-5px}
\subsection{Incremental Setup Details}
After the base model is trained completely we begin with the increment process. 
We formulate this setup as the standard $N$-way $K$-shot few-shot learning i.e. in every increment step, new data come in streams of $N$ new classes with $K$ examples for every class.


In the majority of classification scenarios, the classes are assigned names that comprise words or phrases from natural language. These names frequently carry valuable information that is pertinent to the specific classification problem at hand. For instance, even without any prior knowledge of a white wolf, an average English speaker can make an educated guess that a snow leopard is more similar to a tiger than to a horse. Embeddings of class labels or more elaborate descriptions of classes are often utilized to capture these types of relationships \citep{pennington-etal-2014-glove}.

The information pertaining to class semantics is used to create an enhanced subspace regularizer. This is achieved by promoting new class representations to be in proximity to a convex combination of base classes, with weights assigned based on their semantic similarity.

\begin{table*}[!ht]
\footnotesize
\centering
\caption{Quantitative comparison with state-of-the-art FSCIL approaches on CIFAR-100 dataset in multi-session setting.}
\begin{tabular}{l*{10}{c}}
\toprule
Method/Session No. & 0 & 1 & 2 & 3 & 4 & 5 & 6 & 7 & 8 & Improvement \\
\midrule
GFR \citep{gfr} & 36.34 & 25.21 & 17.34 & 12.41 & 10.22 & 8.76 & 7.34 & 7.01 & 6.22 & +51.35 \\
iCaRL \citep{rebuffi2017icarl} & 64.10 & 53.28 & 41.69 & 34.13 & 27.93 & 25.06 & 20.41 & 15.48 & 13.73 & +43.84 \\
EEIL \citep{eeil} & 64.10 & 53.11 & 43.71 & 35.15 & 28.96 & 24.98 & 21.01 & 17.26 & 15.85 & +41.72 \\
NCM \citep{ncm} & 64.10 & 53.05 & 43.96 & 36.97 & 31.61 & 26.73 & 21.23 & 16.78 & 13.54 & +44.03 \\

\midrule
TOPIC \citep{tao2020few_topic} & 64.10 & 55.88 & 47.07 & 45.16 & 40.11 & 36.38 & 33.96 & 31.55 & 29.37 & +28.2 \\
LEC-Net \citep{lecnet} & 64.10 & 53.23 & 44.19 & 41.87 & 38.54 & 39.54 & 37.34 & 34.73 & 34.73 & +22.84 \\
FSLL \citep{fsll} & 64.10 & 55.85 & 51.71 & 48.59 & 45.34 & 43.25 & 41.52 & 39.81 & 38.16 & +19.41 \\
F2M \citep{f2m} & 64.71 & 62.07 & 59.01 & 55.58 & 52.55 & 49.96 & 48.07 & 46.28 & 44.67 & +12.9 \\
FSIL-GAN \citep{fsil_gan} & 70.14 & 64.36 & 57.21 & 55.21 & 54.34 & 51.89 & 50.12 & 47.91 & 46.61 & +10.96 \\
MetaFSCIL \citep{metafscil} & 74.50 & 70.10 & 66.84 & 62.77 & 59.48 & 56.52 & 54.36 & 52.56 & 49.97 & + 7.6 \\
Data-free Replay \citep{data_free} & 74.40 & 70.20 & 66.54 & 62.51 & 59.71 & 56.58 & 54.52 & 52.39 & 50.14 & +7.43 \\
\midrule
Ours & \textbf{80.24} & \textbf{72.32} & \textbf{69.27} & \textbf{67.14} & \textbf{64.92} & \textbf{65.53} & \textbf{62.52} & \textbf{59.1} & \textbf{57.57} & --- \\
\bottomrule
\end{tabular}
\vspace{-10px}
\label{multi_cifar}
\end{table*}

\begin{table*}[!ht]
\footnotesize
\centering
\caption{Quantitative comparison with state-of-the-art FSCIL approaches on miniImageNet dataset in multi-session setting.}
\begin{tabular}{l*{10}{c}}
\toprule
Method/Session No. & 0 & 1 & 2 & 3 & 4 & 5 & 6 & 7 & 8 & Improvement \\
\midrule
\cite{tao2020few_topic} & 61.31 & 50.09 & 45.17 & 41.16 & 37.48 & 35.52 & 32.19 & 29.46 & 24.42 & +25.28 \\
\cite{cheraghian2021semantic} & 62.00 & 58.00 & 52.00 & 49.00 & 48.00 & 45.00 & 42.00 & 40.00 & 39.00 & +10.7 \\
Subspace reg. \citep{akyurek2021subspace} & 80.37 & 73.76 & 68.36 & 64.07 & 60.36 & 56.27 & 53.10 & 50.45 & 47.55 & +2.15 \\
Data-free Replay \citep{data_free} & 71.84 & 67.12 & 63.21 & 59.77 & 57.01 & 53.95 & 51.55 & 49.52 & 48.21 & +1.49 \\
MetaFSCIL \citep{metafscil} & 72.04 & 67.94 & 63.77 & 60.29 & 57.58 & 55.16 & 52.79 & 50.79 & 49.19 & +0.51 \\
\midrule
Ours & \textbf{81.46} & \textbf{74.08} & \textbf{69.39} & \textbf{65.53} & \textbf{61.75} & \textbf{57.74} & \textbf{53.88} & \textbf{51.41} & \textbf{49.7} & --- \\
\bottomrule
\end{tabular}
\vspace{-15px}
\label{multi_mini}
\end{table*}

\begin{table*}[!ht]
\footnotesize
\centering
\caption{Quantitative comparison with state-of-the-art memory variant FSCIL approaches on miniImageNet dataset in multi-session setting.}
\begin{tabular}{lcccccccccc}
\toprule
Method/Session No. & 0 & 1 & 2 & 3 & 4 & 5 & 6 & 7 & 8 & Improvement \\
\midrule
\cite{chen2020incremental} & 64.77 & 59.87 & 55.93 & 52.62 & 49.88 & 47.55 & 44.83 & 43.14 & 41.84 & +9.71\\

Subspace reg. \citep{akyurek2021subspace} & 80.37 & \textbf{73.92} & \textbf{69.00} & 65.10 & 61.73 & 58.12 & 54.98 & 52.21 & 49.65 & +1.9 \\
 
\midrule
Ours & \textbf{80.98} & 73.37 & 68.31 & \textbf{65.69} & \textbf{62.68} & \textbf{58.89} & \textbf{55.68} & \textbf{53.66} & \textbf{51.55} & --- \\

\bottomrule
\end{tabular}
\vspace{-15px}
\label{multi_mini_memory}
\end{table*}

\begin{table}[!ht]
\footnotesize
\centering
\caption{Quantitative comparison with state-of-the-art memory variant FSCIL approaches on tieredImageNet dataset in single-session setting.}
\begin{tabular}{lcc}
\toprule
Method & Accuracy & $\Delta$ \\
\midrule
Imprinted Networks \citep{imprinted} & $53.87 \pm {\scriptstyle  0.48}$ & \textbf{-17.18}\% \\
LwoF \citep{lwof} & $63.22 \pm {\scriptstyle  0.52}$ & -7.27\% \\
Attention Attractor Networks \citep{ren2019incremental} & $65.52 \pm {\scriptstyle  0.31}$ & -4.48\% \\
XtarNet \citep{yoon2020xtarnet} & $69.58 \pm {\scriptstyle  0.32}$ & -1.79\% \\
 
Subspace Reg. \citep{akyurek2021subspace} & $73.51 \pm {\scriptstyle  0.33}$ & -6.08\%  \\

\midrule
Ours & $\textbf{74.00} \pm {\scriptstyle  0.17}$ & -6.92\%  \\

\bottomrule
\end{tabular}
\vspace{-15px}
\label{tiered_single}
\end{table}

Given a stream of incoming new weight vectors $\eta_{c \in C^{(t)}}$
we optimize the following loss function:
\begin{equation}
L(\eta)= -\alpha||\eta||^{2}  -\beta R_{old}^{(t)}(\eta) -\gamma R_{new}^{(t)}(\eta)
\end{equation}

Here, $R_{old}^{(t)}$ restricts the
extent of fine-tuning of the previously learned classes. $R_{old}^{(t)}$ is given by,
\begin{equation}
R_{old}^{(t)}(\eta)=\sum_{t^{'}<t}\sum_{c\in C^{(t^{'})}} || \eta_{c}^{t^{'}} - \eta_{c} ||^{2} 
\end{equation}

where $\eta_{c}^{t^{'}}$ measures the variable's at the end of session $t^{'}$.

$R_{new}^{(t)}$ provides semantic subspace regularization for novel classes, which is given by,
\begin{equation}
R_{new}^{(t)}(\eta)=\sum_{c\in C^{(t)}} || \eta_{c} - l_{c}||^{2}
\end{equation}
where,
\begin{equation}
l_{c}=\sum_{j \in C^{(0)}} \frac{ exp(e_{j}.e_{c}/ \tau )       }{    \sum_{c^{'} \in C^{(0)}}  exp(e_{c^{'}}.e_{c}/ \tau )   } \eta_{j}
\end{equation}

Here, $\tau$ is a hyper-parameter. Embeddings $e_{c}$ can be derived from standard language models. 

\textbf{Memory Variant}:
In line with previous research conducted by  \cite{chen2020incremental} that focuses on incremental learning while preserving a small ``memory" of past samples denoted as $M$, we investigate an alternative baseline approach. We define the memory at session $t$ as $ M^{(t)} = \bigcup _{(t^{'}<t)} M^{(t^{'})} $ where $M^{(t^{'})}  \subseteq S^{(t^{'})}$ and $|M^{(t^{'})}|=|C^{(t^{'})}|$ In this approach, we adopt a sampling strategy where we select only one example from each previous class, and we continue to reuse the same example in subsequent sessions.

\subsection{Insights}
Our core novelty lies in bridging the domain gap between language and text to make the base model adapt to additional information coming from the text semantics early on in the training pipeline. This is achieved by adding an additional relational term in the loss function capturing the text-to-image relationship. But this increased degree of freedom obtained by 
the model may lead to overfitting without a lot of training examples. To tackle this we go for the K-NN based Laplacian regularizer as described in previous sections. We optimally set the value of K to balance the trade-off between overfitting due to additional dimension and no semantic information at all.

For incremental session training, the semantic subspace regularization approach works fine to take care of catastrophic forgetting.


\begin{table*}[ht]
\footnotesize
\centering
\caption{Importance of language regularizer in CIFAR-100.}
\begin{tabular}{lccccccccc}
\toprule
Config. / Session No. & 0 & 1 & 2 & 3 & 4 & 5 & 6 & 7 & 8 \\

\midrule
w/o Lang. Reg. & 70.14 & 64.36 & 57.21 & 55.21 & 54.34 & 51.89 & 50.12 & 47.91 & 46.61 \\
\midrule
w/ Lang. Reg.  & \textbf{80.24} & \textbf{72.32} & \textbf{69.27} & \textbf{67.14} & \textbf{64.92} & \textbf{65.53} & \textbf{62.52} & \textbf{59.1} & \textbf{57.57} \\
 
\bottomrule
\end{tabular}
\vspace{-15px}
\label{abl_0}
\end{table*}

\begin{table*}[!ht]
\footnotesize
\centering
\caption{Effect of different variants of language regularizer in CIFAR-100.}
\begin{tabular}{lccccccccc}
\toprule
Loss Fn. / Session No. & 0 & 1 & 2 & 3 & 4 & 5 & 6 & 7 & 8 \\
\midrule
Euclidean distance & 80.90 & \textbf{75.51} & 70.11 & \textbf{69.64} & 63.84 & \textbf{66.67} & \textbf{63.46} & 58.69 & 55.62 \\
\midrule
Cosine similarity & 80.09 & 74.22 & 69.14 & 69.13 & 62.49 & 66.12 & 63.25 & \textbf{59.52} & 56.72  \\
 \midrule
Top-K Cosine similarity  & \textbf{80.24} & 72.32 & \textbf{69.27} & 67.14 & \textbf{64.92} & 65.53 & 62.52 & 59.1 & \textbf{57.57} \\
 
\bottomrule
\end{tabular}
\vspace{-15px}
\label{abl_1}
\end{table*}

\begin{table*}[!ht]
\footnotesize
\centering
\caption{Effect of different semantic representations used in Language Regularizer for CIFAR-100.}
\begin{tabular}{lccccccccc}
\toprule
Representation / Session No. & 0 & 1 & 2 & 3 & 4 & 5 & 6 & 7 & 8 \\
\midrule
GloVe & 79.87 & 72.72 & 69.63 & 66.96 & 64.48 & 64.27 & 61.17 & 57.56 & 56.89 \\
\midrule
fastText & 80.21 & \textbf{72.90} & \textbf{69.77} & \textbf{67.14} & \textbf{64.95} & 63.82 & 61.46 & 57.92 & 57.33 \\
\midrule
CLIP  & \textbf{80.24} & 72.32 & 69.27 & \textbf{67.14} & 64.92 & \textbf{65.53} & \textbf{62.52} & \textbf{59.1} & \textbf{57.57} \\
\midrule
\end{tabular}
\vspace{-15px}
\label{tab_embedding}
\end{table*}

\begin{table*}[!ht]
\footnotesize
\centering
\caption{Effect of different values of K in Language Regularizer for CIFAR-100.}
\begin{tabular}{lccccccccc}
\toprule
K / Session No. & 0 & 1 & 2 & 3 & 4 & 5 & 6 & 7 & 8 \\
\midrule
3 & 79.74 & \textbf{72.87} & 69.37 & 66.73 & 64.34 & 64.14 & 61.20 & 57.52 & 57.29 \\
\midrule
5  & \textbf{80.24} & 72.32 & 69.27 & \textbf{67.14} & \textbf{64.92} & \textbf{65.53} & \textbf{62.52} & \textbf{59.1} & \textbf{57.57} \\
\midrule
10 & 80.07 & 72.81 & \textbf{69.72} & 66.96 & 64.39 & 63.89 & 61.25 & 57.58 & 57.09 \\
\midrule
\end{tabular}
\vspace{-15px}
\label{tab_k}
\end{table*}

\section{Experiments and Results}
\subsection{Datasets}
We perform our experiments and present results on three datasets: miniImageNet, tieredImageNet and CIFAR-100. We follow the standard protocols for FSCIL evaluation. Both miniImageNet and CIFAR-100 consist of a total of 100 classes out of which 60 classes are reserved for base session and the remaining 40 classes come in the form of 8 streams, each stream bearing 5 novel classes.
The few-shot learning during incremental episodes is 5-way and 5-shot in nature.

The tieredImagenet has a total of 608 classes out of which 351 classes form the base class set and 160 classes are introduced in a single incremental session. The few-shot setup used here is 160-way and 5-shot.

\subsection{Evaluation}
Our experiments are designed to assess the learning of new classes and the retention of base classes in a classifier initially trained on a set of base classes. To examine the effectiveness of our method, we evaluate it using two different experimental paradigms commonly employed in previous studies.

The first paradigm involves a multi-session experiment, where new classes are continuously introduced over multiple sessions, requiring the classifier to be updated repeatedly. For this, we have used miniImageNet and CIFAR-100 datasets. The second paradigm is a single-session setup, where new classes are presented only once during training. We utilize tieredImageNet for the single-session paradigm. For training all the models, we use Stochastic Gradient Descent (SGD) as our optimizer.


\textbf{Multi-session results} We present session-wise accuracy of our method along with multiple state-of-the-art methods on CIFAR-100 and miniImageNet dataset in Table-\ref{multi_cifar} and Table-\ref{multi_mini}. We can see that our method performs better than other approaches.

In Table-\ref{multi_mini_memory}, we have reported performance of our memory variant model on miniImageNet dataset with other state-of-the-art memory variant approaches, where we see that our method outperforms previous methods.

\textbf{Single-session results} We have reported performance of our framework and other state-of-the-art methods on single-session setting in Table-\ref{tiered_single}. Along with accuracy values, we have also reported $\Delta$ metric \citep{ren2019incremental}, which is the difference between
individual accuracies and joint accuracies of both base and novel samples averaged. It can be seen that our method achieves better accuracy than other state-of-the-art methods while achieving decent $\Delta$ score.

\subsection{Ablation Experiments}

\textbf{Significance of language regularizer:} To analyze the importance of language regularizer, we have trained a model without using it and compared the performance against our model trained with language regularizer in Table-\ref{abl_0}. It is evident that language regularizer vastly improves the accuracy of the model with respect to the model trained without language regularizer.

\textbf{Effect of different similarity/distance measure in language regularizer loss:} We have experimented with various similarity/distance measures to compute $\lambda$ and $\mu$ in language regularizer: (a) Euclidean distance, (b) Cosine similarity, (c) Top-K Cosine similarity. For (a) and (b), we consider all base classes, whereas we consider only top-K neighbors as shown in Equ-7 for (c).

In Table-\ref{abl_1}, we can see that our framework performs best at the end of all the training sessions when ``top-K cosine similarity" is used in language regularizer.

\textbf{Effect of different semantic representations} 
We have tried three different class label representations in our language regularizer: GloVe \citep{pennington-etal-2014-glove}, fastText \citep{fasttext} and CLIP \citep{clip}. 

For extracting semantic embeddings using GloVe or fastText, we simply extract word embeddings for single-word class labels. For multi-word class labels, we first extract embeddings for each of the word in class label and compute the average of individual embeddings. 

For obtaining class semantic representation using CLIP, we utilize the text encoder of CLIP.  two different prompts i.e. ``A photo of \texttt{class\_label}" and ``A good photo of \texttt{class\_label}". We prompt the text encoder of CLIP with these two prompts and compute the average of the output text representations to get the final representation.

We report the quantitative comparison between two representations in Table-\ref{tab_embedding}. We observe that CLIP representations work better in most training episodes. It can be attributed to the fact that GloVe and fastText embeddings are trained on text-only data whereas CLIP is trained on image-text data hence providing rich semantic representation which is more useful in FSCIL.



\section{Conclusion}
In this study, we introduced a novel framework for few-shot class incremental learning (FSCIL) that leverages language regularizer and subspace regularizer to seamlessly integrate new classes with limited data while preserving base class performance. Our approach reduces the semantic gap between visual and textual (class labels) modalities in base model training by introducing a cross-domain graph Laplacian regularizer. The effectiveness of our framework was demonstrated through comprehensive experiments on various FSCIL benchmarks in both single-session and multi-session settings, where it achieved state-of-the-art performance. 
We have used an average of two prompt embeddings for obtaining semantic embeddings for class labels, however, future works can explore more on designing better prompts and ensembling these prompts. Finally, we have used CLIP as Vision-Language model in this paper, although similar models such as BLIP \citep{li2022blip} or FLIP \citep{li2023scaling} can also be used. 

\vspace{-15px}

\bibliographystyle{model5-names}
\bibliography{refs}

\end{document}